\documentclass[conference]{IEEEtran}
\IEEEoverridecommandlockouts

\usepackage{cite}
\usepackage{amsmath,amssymb,amsfonts}
\usepackage{algorithmic}
\usepackage{graphicx}
\usepackage{textcomp}
\usepackage{xcolor}
\usepackage{caption}
\usepackage{subcaption}
\usepackage{booktabs} 
\usepackage{hyperref}
\def\BibTeX{{\rm B\kern-.05em{\sc i\kern-.025em b}\kern-.08em
    T\kern-.1667em\lower.7ex\hbox{E}\kern-.125emX}}
\begin{document}

\title{Hybrid Quantum-Classical Corrective Diffusion Modeling for Meteorological Downscaling\\

\thanks{This work is part of the ``Quantum Readiness for Earth Observation'' (QR4EO) project, which is funded by the European Space Agency (ESA) $\Phi$-lab under contract No. 4000147474/25/I-DT. The authors gratefully acknowledge the Gauss Centre for Supercomputing e.V. (www.gausscentre.eu) for funding this project by providing computing time on the GCS Supercomputer JUWELS at J\"ulich Supercomputing Centre (JSC). This work is partially supported by the Icelandic Centre for Research, project \textit{"Hybrid Quantum-Classical Workflows for EO"}, RANNÍS grant number: 2511078-051 (\url{http://en.rannis.is/}).

This work has been submitted to the IEEE for possible publication. Copyright may be transferred without notice, after which this version may no longer be accessible.
}
}

\author{
\IEEEauthorblockN{Rui Wang\IEEEauthorrefmark{1}\IEEEauthorrefmark{2}, 
Edoardo Pasetto\IEEEauthorrefmark{1}\IEEEauthorrefmark{3}, 
Amer Delilbasic\IEEEauthorrefmark{1}\IEEEauthorrefmark{2},  
Morris Riedel\IEEEauthorrefmark{1}\IEEEauthorrefmark{2},
Kristel Michielsen\IEEEauthorrefmark{1}\IEEEauthorrefmark{4},  
Gabriele Cavallaro\IEEEauthorrefmark{1}\IEEEauthorrefmark{2}}
\IEEEauthorblockA{\IEEEauthorrefmark{1}Jülich Supercomputing Centre, Forschungszentrum Jülich, Jülich, Germany \\
Email:\{r.wang, e.pasetto, a.delilbasic, m.riedel, k.michielsen, g.cavallaro\}@fz-juelich.de}
\IEEEauthorblockA{\IEEEauthorrefmark{2}University of Iceland, Reykjavík, Iceland}
\IEEEauthorblockA{\IEEEauthorrefmark{3}RWTH Aachen University, Aachen, Germany}
\IEEEauthorblockA{\IEEEauthorrefmark{4}University of Cologne, Cologne, Germany}

}

\maketitle

\begin{abstract}
Statistical downscaling is a crucial component of the weather modelling field, where high-resolution outputs must be reconstructed from coarse-resolution inputs with the full cost of dynamical refinement. In this work, we investigate a hybrid quantum-classical corrective diffusion model for probabilistic statistical downscaling of weather fields. The proposed model inserts variational quantum circuit (VQC) layers into the most compressed bottleneck of the diffusion UNet while leaving the regression branch fully classical. This placement tests whether quantum circuits can act as compact nonlinear feature maps for latent-channel mixing. We evaluate intra-channel and cross-channel ans\"atze on 10\,m wind components. On the 2020 validation set, the hybrid models remain stable, preserve the large-scale spatial organisation of the generated wind fields, and improve both mean absolute error (MAE) and continuous ranked probability score (CRPS) relative to a classical corrective diffusion model in several configurations. Structural diagnostics further show that the hybrid variants preserve kinetic-energy spectra and windspeed distributions similar to its classical counterpart while producing controlled changes in tail behavior, extreme-windspeed localization, and joint wind field components structure. Backend studies on the 2020 validation set show negligible impact from simulated device noise at the tested circuit scale, whereas real-hardware deployment remains limited by qubit availability and execution fidelity. Furthermore, the 2021 out-of-distribution test shows that these in-distribution gains do not transfer uniformly under temporal shift, revealing a generalization gap that motivates future mitigation through stabilization and regularization. These results show that bottleneck-level quantum hybridization can make a nontrivial contribution to weather statistical downscaling, while also highlighting that circuit scale and hardware deployment remain key limiting factors.
\end{abstract}


\section{Introduction}

\subsection{Overview of statistical downscaling}
Statistical downscaling \cite{benestad2008, stoner2013, Thrasher2022, SACHINDRA2018240, Sachindra2011} is a computational technique used to enhance the spatial resolution of coarse geospatial data, making it usable for local-scale applications. Data from Global Climate Models (GCM) is often generated at low resolutions that 
is not sufficient to be employed in many applications at a local scale \cite{Iorio2004, Sachindra2015, gutowski2020}.
The method works by learning the statistical relationships between large-scale, low-resolution data (e.g., from a global climate model) and historical, fine-scale measurements of a target variable (e.g., local precipitation). Once this relationship is established from historical data, it can be applied as a predictive model to new, unseen coarse data to generate a corresponding high-resolution output.

Statistical downscaling is a key enabling technology for weather and climate digital twins, where it can refine forecasts and simulations to decision-relevant scales without prohibitive computational costs \cite{Rashid2015, Sachindra2015}. Its performance can be measured by system-level KPIs like speed and accuracy, and application-level KPIs that assess physical realism and usability for specific weather variables and operational integration.

Several machine learning techniques have been applied in the literature to statistical downscaling tasks. Some examples include generative adversarial networks \cite{Stengel2020, Leinonen2021, Vosper2023}, vision transformer-based models \cite{nguyen2023climax} and diffusion models \cite{molinaro2026universal, hatanaka2023diffusion}. In this work, we focus on diffusion models, whose stochastic nature makes them suitable to tackle the atmospheric downscaling problem at a high spatial resolution \cite{Selz2015}.

\subsection{Overview of classical diffusion models}
Diffusion models are a popular class of generative models whose functioning revolves around gradually deteriorating the input data through the pixel-wise addition of Gaussian noise and subsequently learning how to return to the original image by inverting the noise process \cite{ho2020denoising, yang2023diffusion}. Once the \textit{denoising} process is learned, it is possible to generate new images by denoising some sample instances of Gaussian noise that are easy to sample.
The noise corruption process employed in diffusion models happens in a "Markovian" way. Starting from the input image $\mathbf{x}_0$, a sequence of T images is sampled according to the conditional distributions $q(\mathbf{x}_t | \mathbf{x}_{t-1})$ that are defined based on a given noise schedule. At the end of the data corruption process, the latent image representation $\mathbf{x}_T$ is not recognizable from pure Gaussian noise. During the training phase, the goal is to train a deep learning model that, given a sample of Gaussian noise, can denoise it through the subsequent sampling from the conditional distributions $p_\theta(\mathbf{x}_{t-1} | \mathbf{x}_t)$ up until the original image space. In this way, the diffusion model learns a stochastic mapping from a space of Gaussian noise samples to the space of the input images.

\section{Methods}

\subsection{Diffusion models for statistical downscaling}
\begin{figure}[h!]
  \centering
  \includegraphics[width=0.99\linewidth,
    clip]{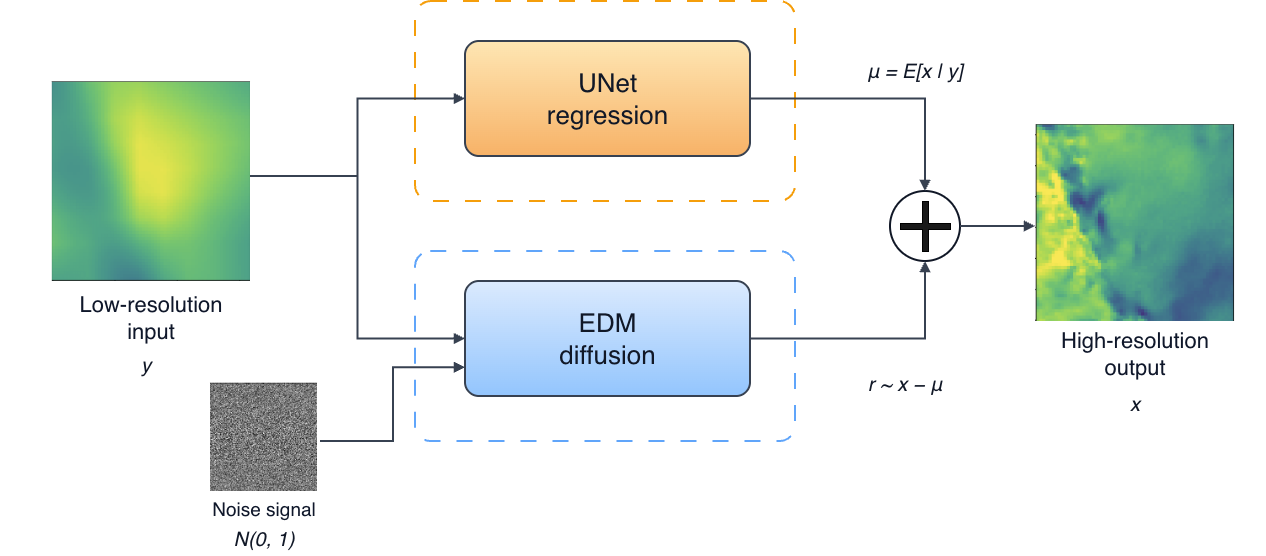}
  \caption{High-level schematic of the CorrDiff workflow proposed by Mardani et al. \cite{mardani2025residual}. A low-resolution input is first mapped by a deterministic UNet regressor. The low-resolution input, noise signal, and regressor output are then passed to a diffusion model that predicts a residual correction. The final prediction is obtained by adding the residual to the regressor output.}
  \label{fig:corrdiff}
\end{figure}

The implementation developed in this work is based upon the CorrDiff \cite{mardani2025residual} super-resolution diffusion model. The workflow for training and sampling CorrDiff for generative downscaling is shown in fig. \ref{fig:corrdiff}.
The objective of CorrDiff is, given a low-resolution input image \textbf{y} representing a specific meteorological physical quantity of interest, to provide as output a better-resolution image $\mathbf{x}$. This process is carried out in a probabilistic way by finding the conditional distribution $p(\mathbf{x} | \mathbf{y})$ and sampling from it. The output prediction $\mathbf{x}$ is found in a 2-step process: in the first step, a deterministic regression UNet \cite{ronneberger2015u} predicts the mean $\mathbb{E}[\mathbf{x} | \mathbf{y}]$, while in the second step, the diffusion model is trained to stochastically generate the residual $\mathbf{r} =\mathbf{x} - \mathbb{E}[\mathbf{x} | \mathbf{y}]$. At inference time, the final prediction $\mathbf{x}$ is reconstructed as $\mathbf{x}=\mathbb{E}[\mathbf{x} | \mathbf{y}]+\mathbf{r}$. 

In this work, we study probabilistic statistical downscaling of coarse-resolution weather fields using a hybrid quantum-classical extension of CorrDiff, where the quantum part is derived from the previous work of the quantum hybrid diffusion model \cite{de2024towards}. In our approach, the regression UNet remains fully classical. Quantum-classical hybridization is introduced only in the diffusion branch, where selected convolutional layers inside the bottleneck of the diffusion UNet are replaced by hybrid quantum-classical layers.

\subsection{Hybrid quantum-classical replacement of convolutional layers}

Hybrid quantum processing is introduced inside the UNet block by replacing the classical convolutional layers with a customized PyTorch class at the bottleneck stage of the network. Each hybrid class processes only a subset of the channels through a quantum layer, while a classical convolution processes the remaining channels. The outputs of the quantum and classical branches are then concatenated along the channel dimension. Apart from this substitution, the overall structure of the original UNet block is preserved, including embedding injection, residual connections, positional encoding, and the batch-wise processing scheme.

\begin{figure}[h!]
  \centering
  \includegraphics[width=0.8\linewidth]{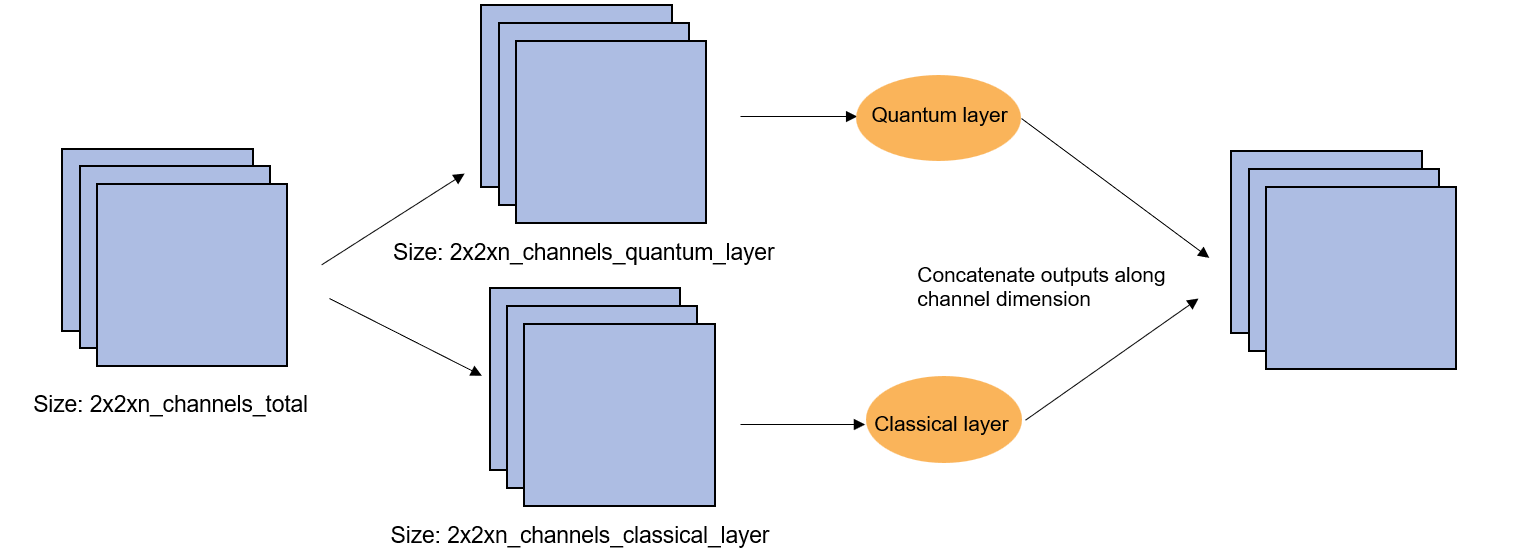}
  \caption{Processing scheme of the customized PyTorch class to handle the hybridization of the quantum and the classical layers. The input feature map is split along the channel dimension: the first $n\_channels\_quantum\_layer := N//4$ (with N being the number of qubits of the ansatz) channels are processed by the quantum channel layer, while the remaining channels are passed through a classical convolutional layer. The outputs of the quantum and classical branches are then concatenated along the channel dimension. \label{fig:hybridconvverrtex}}
\end{figure}

We target the UNet bottleneck because it is the most compressed and semantically dense stage of the network. At this point, the main modeling problem is the transformation of condensed latent channels and the modeling of their joint dependencies rather than local texture extraction. A variational quantum circuit is therefore used here as an alternative nonlinear feature map acting on compact latent representations. Entangling operations provide a natural mechanism for constructing coupled transformations across latent channels, while the small spatial size of the bottleneck keeps the required encoding scale compatible with current quantum hardware.

In the original CorrDiff configuration, the bottleneck resolution is $16\times16$. In contrast, our modified diffusion encoder is extended until a $2\times2$ bottleneck is reached. This design reduces the dimensionality of the latent representation, thereby making bottleneck channels compatible with low-qubit quantum encoding, while still applying the quantum layer to a feature space that contains highly condensed global information.

\subsection{Variational quantum circuit ans\"atze} \label{sec:ansatz}
\begin{figure*}[h!]
    \centering
    \captionsetup{justification=centering}
    \begin{subfigure}[t]{0.34\textwidth}
        \centering
        \includegraphics[width=\textwidth]{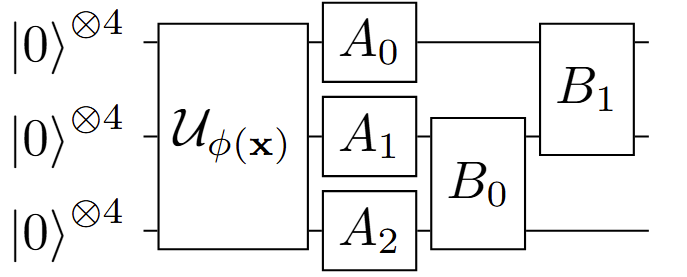}
        \caption{\label{fig:hqconv_subfigure} Overview of the HQConv ansatz used in this work. The circuit structure is based on the HQConv ansatz of \cite{de2024towards}, with a modified encoding block $\mathcal{U}_{\phi(\mathbf{x})}$ that applies Hadamard gates followed by \(R_Z\)-based data encoding.}
    \end{subfigure}
    \begin{subfigure}[t]{0.25\textwidth}
        \centering
        \includegraphics[width=\textwidth]{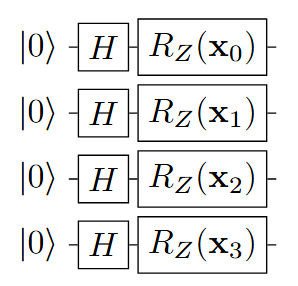}
        \caption{\label{fig:encoding_block} Structure of the encoding block $\mathcal{U}_{\phi(\mathbf{x})}$ used in the adopted ansatz.}
    \end{subfigure}
    \begin{subfigure}[t]{0.24\textwidth}
        \centering
        \includegraphics[width=\textwidth]{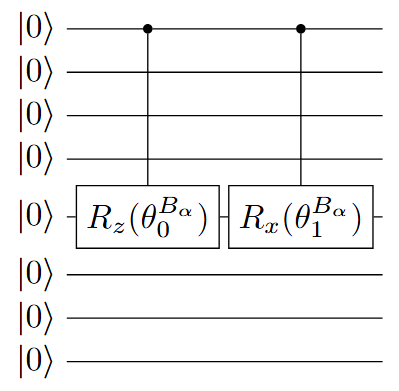}
        \caption{\label{fig:b_block} Structure of block B in the HQConv ansatz. This block introduces correlations across the qubits associated with different input channels.}
    \end{subfigure}
    \begin{subfigure}[t]{0.85\textwidth}
        \centering
        \includegraphics[width=\textwidth]{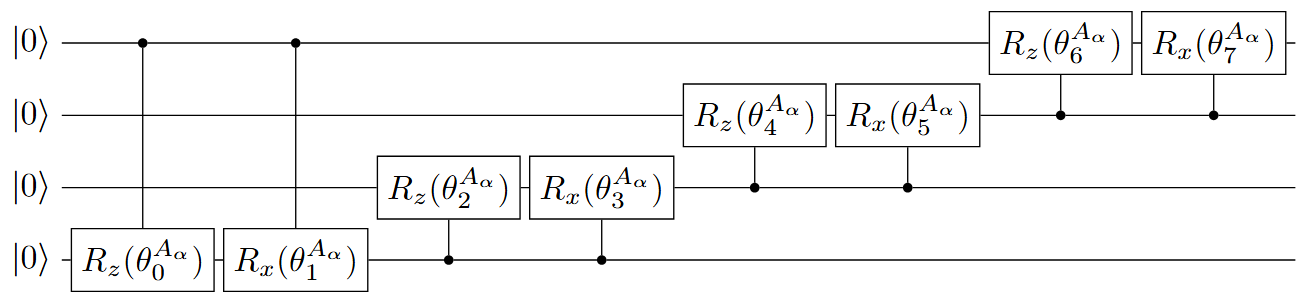}
        \caption{\label{fig:a_block} Structure of block A in the HQConv ansatz. This block introduces correlations among the qubits encoding the features of a single input channel.}
    \end{subfigure}
    \caption{\label{fig:diffusion_ansatz} Overview of the HQConv ansatz. Panels (a)--(d) show the full circuit structure from \cite{de2024towards}, the encoding block, and the constituent A and B blocks. The HQConv ansatz uses 12 qubits and can therefore process three channels of a \(2 \times 2\) input. In our implementation, this construction is generalized to any number of qubits that is a multiple of four, so that the number of processed channels is \(n_{\mathrm{qubits}}/4\). For the special case of four qubits, corresponding to a single input channel, only block A is applied, since block B requires at least two channels.}
\end{figure*}

With the routing of bottleneck features through the quantum branch in place, the remaining design choice is the circuit ansatz that specifies how the selected latent channels are encoded and transformed. In our setting, the ansatz must remain compatible with compact bottleneck encoding while providing a structured mechanism for feature mixing across the latent representation.
We therefore adopt a variational circuit that is inspired by the HQConv ansatz from De Falco et al. \cite{de2024towards}. The circuit contains three conceptual components: an encoding block, which maps classical inputs to qubit rotations (fig. \ref{fig:encoding_block}), Block A, which introduces intra-channel correlations (fig. \ref{fig:a_block}), Block B, which introduces cross-channel correlations (fig. \ref{fig:b_block}).
In our work, multiple ansatz variants are considered:
\begin{itemize}
    \item the full HQConv-inspired ansatz (A+B),
    \item a reduced Block-B-only ansatz,
    \item a 4 qubit Block-A-only ansatz used for real-hardware deployment.
\end{itemize}

The full A+B and Block-B-only ans\"atze are used in the methodology and validation stages. The representative wind-map comparisons reported for these stages use the Block-B-only variant, which achieved the lowest RMSE among the tested configurations. For real-hardware deployment, the ansatz is reduced to a 4-qubit Block-A-only circuit to match the hardware-specific qubit constraints of the target device. In this setting, the design emphasizes intra-channel information mixing within the compact latent representation while remaining executable on the available quantum processor.

\subsection{Channel allocation and circuit multiplicity}

Given this bottleneck-level routing scheme, the hybrid model applies quantum processing to a subset of the 128 bottleneck channels, since each quantum circuit operates on only a small fraction of the latent representation. The exact number of quantum-processed channels depends not only on the number of circuits, but also on the ansatz variant.
For the full A+B and Block-B-only ans\"atze used in the methodology and validation stages, the main configurations are:
\begin{itemize}
    \item \texttt{n\_circuits = 1}, corresponding to quantum processing of 3 bottleneck channels,
    \item \texttt{n\_circuits = 3}, corresponding to quantum processing of 9 bottleneck channels.
\end{itemize}

For the Block-A-only ansatz used in real-hardware deployment, the channel allocation changes because each circuit is restricted to a 4 qubit configuration. In this case:
\begin{itemize}
    \item \texttt{n\_circuits = 1} corresponds to quantum processing of 1 bottleneck channel,
    \item \texttt{n\_circuits = 3} corresponds to quantum processing of 3 bottleneck channels.
\end{itemize}

This distinction is important because the same \texttt{n\_circuits} value does not correspond to the same number of quantum-processed channels across all ansatz variants. In every case, the remaining bottleneck channels are processed classically, so the hybrid model remains a localized modification of an otherwise classical diffusion architecture.


\subsection{Implementation and execution pipeline}

The hybrid architecture was developed within a PyTorch workflow, with the quantum layers realized through PennyLane \cite{bergholm2018pennylane} and integrated into the network via \texttt{qml.qnn.TorchLayer} on the \texttt{default.qubit} simulator. In this setup, each forward pass of the quantum component is evaluated as a full statevector simulation on the CPU. The simulation-based training workflow was executed on a JUWELS node.\footnote{\url{https://www.fz-juelich.de/en/ias/jsc/systems/supercomputers/juwels}}

At the level of the hybrid model, each quantum layer implements a function that returns expectation values of selected observables with respect to the quantum state obtained by applying the parameterized circuit to the reference initial state $|0^n\rangle$. Gradients with respect to the circuit parameters are computed via the parameter-shift rule \cite{schuld2019evaluating}, making the quantum layer differentiable and therefore compatible with end-to-end backpropagation.



For real-hardware deployment, the workflow was further adapted to the target IQM system. This stage imposes hardware specific constraints that do not apply during training or backend emulation, most notably in the number of available qubits, and therefore requires the use of a 4 qubit Block-A-only ansatz. 
Deployment side layer implementations were introduced to preserve the same high level interface as the original quantum modules while accommodating hardware execution and batched job submission. The hardware executed model should therefore be understood as a constrained realization of the same bottleneck hybridization principle, rather than as a one to one reproduction of every ansatz used during validation.

\subsection{Benchmarking setting}

Benchmarking is performed on the HRRR-mini dataset, a reduced weather dataset derived from the HRRR CONUS domain and collected over the period 2018–2021. It covers the continental United States and surrounding regions and provides a computationally manageable testbed for weather downscaling experiments while retaining the structure of a realistic mesoscale forecasting setting.\footnote{\url{https://catalog.ngc.nvidia.com/orgs/nvidia/teams/modulus/resources/modulus_datasets-hrrr_mini?version=1}} The present study focuses on two output variables: 10-meter zonal wind (u10m) and 10-meter meridional wind (v10m). This reduced variable set keeps the hybrid experiments tractable while preserving a meaningful atmospheric prediction task.

\subsection{Evaluation metrics and structural diagnostics} \label{sec:evaluation_metrics}

We evaluate model performance using root mean square error (RMSE), mean absolute error (MAE), and continuous ranked probability score (CRPS), as these metrics capture complementary aspects of prediction quality in a probabilistic downscaling setting. 

RMSE is used for representative spatial map comparisons, where larger local deviations should be penalized more strongly than small pointwise errors. In this context, RMSE is a useful metric for summarizing visible deviations in reconstructed wind-speed maps, where localized spatial errors and misrepresented high intensity structures should contribute more strongly to the overall score than small background differences.

MAE is used to assess pointwise error in the generated fields in a way that is directly interpretable in the physical units of the target variables. Compared with squared error measures, MAE provides a more linear summary of deterministic prediction error and is therefore useful for tracking whether the hybrid model improves average agreement with the reference fields across sampled timestamps.

CRPS is used to evaluate the probabilistic predictions produced by the diffusion model \cite{gneiting2007strictly}. As a strictly proper scoring rule for univariate probabilistic forecasts, CRPS assesses the full predictive distribution rather than only a single point estimate. It is especially appropriate here because the diffusion model generates stochastic predictions, and the quality of the forecast must therefore be judged not only by accuracy, but also by how well the predictive distribution aligns with the realized target. In addition, CRPS generalizes the absolute error and reduces to it in the deterministic limit, which makes it a natural probabilistic counterpart to MAE.

Because pointwise scores do not fully characterize the realism of downscaled atmospheric fields, we additionally use structural diagnostics for probing multiscale variability, distributional realism, extreme-event localization, and multivariate consistency. Directional 10\,m kinetic-energy spectra are used to assess whether the generated wind fields reproduce scale-dependent atmospheric variance, a standard diagnostic in numerical weather prediction and high-resolution model evaluation \cite{skamarock2014atmospheric,mardani2025residual}. Windspeed log-PDFs are used to evaluate marginal distributional realism and tail behaviour, following the CorrDiff-style analysis of whether corrective diffusion restores variability suppressed by deterministic regression models \cite{mardani2025residual}.

Extreme-event localization is evaluated with the Fractional Skill Score (FSS), a neighborhood-based metric for threshold exceedances \cite{roberts2008scale}. We use FSS for high-percentile windspeed events to test whether extreme wind regions are placed correctly up to a prescribed spatial tolerance. Finally, the joint $(u10m,v10m)$ density is used as a diagnostic of multivariate wind consistency, since matching univariate marginals alone does not guarantee preservation of inter-variable dependence~\cite{cannon2016multivariate,wang2024multivariate}.

Taken together, these metrics and diagnostics allow us to evaluate spatial reconstruction quality, deterministic pointwise accuracy, and probabilistic forecast quality, multiscale spectral realism, distributional tail behaviour, extreme-event localization, and multivariate wind consistency within a common benchmarking framework. Execution time, parameter count, and qubit count are treated separately as implementation and computational descriptors rather than as primary predictive metrics.

\section{Results and discussion}

This section evaluates the hybrid quantum-classical CorrDiff model from five complementary perspectives. It first examines whether the hybrid bottleneck in the diffusion UNet backbone preserves the qualitative structure of the generated downscaled wind fields. Second, it quantifies predictive performance on the 2020 in-distribution (ID) evaluation set using MAE and CRPS and compares the ans\"atze discussed in Section \ref{sec:ansatz} and channel count configurations against the classical CorrDiff baseline. Third, pointwise metrics are complemented with structure-oriented diagnostics, including kinetic-energy spectra, windspeed distributions, extreme-event localization, and joint $(u,v)$ statistics. Fourth, it separates simulation-based validation from deployment-oriented execution by comparing noiseless inference, noise-aware emulation, and real-hardware deployment under strict qubit constraints. Finally, we utilize the 2021 out-of-distribution (OOD) results to assess the robustness of ID improvements against temporal distribution shifts. The resulting generalization gap directly motivates the need for mitigation strategies.

\subsection{Qualitative downscaling behaviour}

\begin{figure*}[ht]
  \centering
  \includegraphics[width=\textwidth]{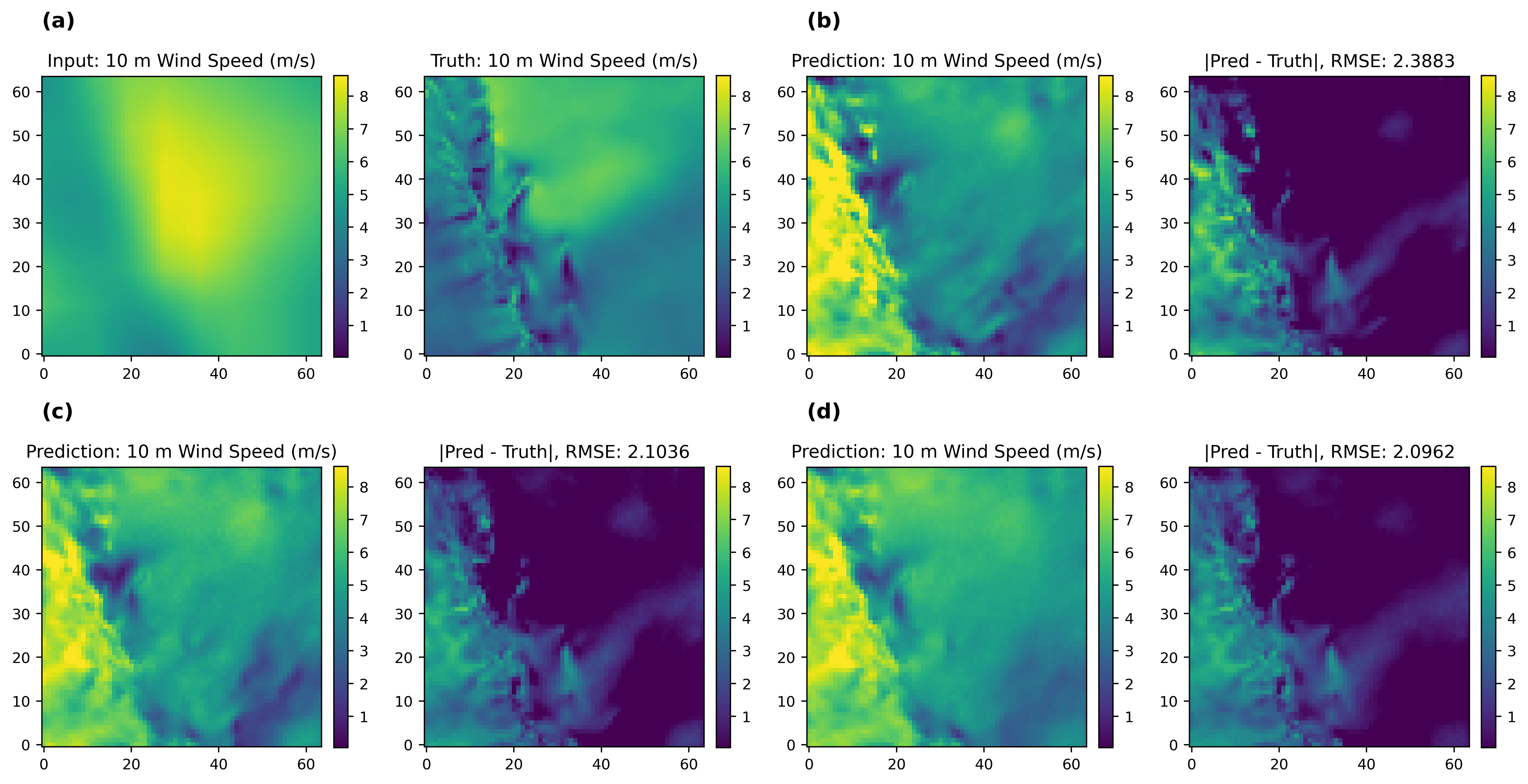}
  \caption{Representative 10\,m wind-speed comparisons for the classical and hybrid CorrDiff models under matched bottleneck resolution. The conditioning input and truth are shown once for reference (a), while the CorrDiff rows display the prediction and residual fields for fully classical (b), Block-B-only ansatz with \texttt{n\_circuit=1} ($n_{ch}=3$) (c), and Block-B-only ansatz with \texttt{n\_circuit=3} ($n_{ch}=9$) (d). The qualitative comparison indicates that the hybrid bottleneck preserves the coarse flow organization while improving local structure relative to a classical model with the same bottleneck resolution.}
  \label{fig:downscaling_test}
\end{figure*}

This subsection examines whether the hybrid bottleneck modification changes the qualitative structure of the generated wind fields and whether the gains can be attributed solely to a deeper classical encoder. The representative wind-speed field maps consist of the classical CorrDiff model, modified to use the same bottleneck resolution as the hybrid architecture, against two representative hybrid models with \texttt{n\_circuit=1} and \texttt{n\_circuit=3} (figure~\ref{fig:downscaling_test}). To avoid repetition of the same conditioning fields, the input and truth maps are shown once (fig.~\ref{fig:downscaling_test}(a)), while the classical CorrDiff (fig.~\ref{fig:downscaling_test}(b)) and the hybrid rows \texttt{n\_circuit=1} (fig.~\ref{fig:downscaling_test}(c)) and \texttt{n\_circuit=3} (fig.~\ref{fig:downscaling_test}(d)) emphasize only the prediction and residual structure. The figure shows that the representative hybrid models remain consistent with the classical CorrDiff baseline and preserve the dominant large-scale wind flow. At the same time, hybrid predictions recover sharper local structures than the classical reference, while also reducing the residual magnitude.
Since the classical CorrDiff baseline already uses the same deeper bottleneck geometry ($2\times2$ bottleneck resolution) as the hybrid model, these improvements cannot be explained solely by additional downsampling depth or by a more compressed latent representation. Instead, the results indicate that the HQConv-based bottleneck contributes nontrivially to the generated field beyond what is obtained by a purely classical encoder with matched resolution.

\subsection{Quantitative validation and ansatz dependence on the 2020 evaluation set}

This subsection quantifies predictive skill on the 2020 evaluation set using MAE and CRPS, and relates the observed performance differences to ansatz structure and quantum channel count. Following the CorrDiff evaluation logic, MAE assesses deterministic field accuracy using the ensemble mean, while CRPS evaluates the probabilistic quality of the predictive distribution generated by the diffusion model. The  main hybrid configurations are summarized through mean MAE, mean CRPS, and total per-timestep hybrid model win counts in table \ref{tab:mae_crps_summary}.

Across the tested variants, the hybrid and classical trajectories remain close throughout the evaluation period, indicating that the insertion of the quantum bottleneck does not produce large systematic instabilities. The quantitative performance differences instead appear as moderate and configuration-dependent in average behaviour. In the smaller-channel regimes, the full A+B and A-only variants remain stable but provide limited gains. The full A+B configuration with $n_{ch}=3$ slightly degrades all four mean metrics relative to the classical baseline, suggesting that combining spatial and cross-channel transformations is not beneficial when only a small number of bottleneck channels is processed quantum-mechanically. The A-only variants, which act within individual latent channels, also remain close to the baseline for $n_{ch}=1$ and $n_{ch}=3$, with win rates below 50\%. This indicates that local spatial processing within isolated channels is not sufficient at low channel count to consistently improve the generated wind fields.

The exception in the smaller-channel setting is the B-only configuration with $n_{ch}=3$. By focusing on interactions across latent channels, this variant slightly improves both MAE components relative to the classical baseline and reaches a 50.0\% win rate. Although the improvement is modest, it suggests that even a small amount of cross-channel bottleneck mixing is more useful than isolated within-channel processing in the low-capacity regime. This observation also supports that increasing the number of quantum-processed channels is important.

Stronger behavior emerges when the number of quantum-processed bottleneck channels is increased to $n_{ch}=9$. In this regime, the ansatz comparison becomes more differentiated rather than being dominated by a single variant. The A-only model achieves the best $u10m$ performance, with the lowest MAE-$u10m$ and CRPS-$u10m$ among all tested configurations. This suggests that repeated quantum processing within individual latent channels can be beneficial for the more coherent large-scale structure of the zonal wind component once enough channels are included. The full A+B model achieves the best $v10m$ performance, yielding the lowest MAE-$v10m$ and CRPS-$v10m$, which indicates that the combined spatial and cross-channel circuit is more useful for the component whose errors are more sensitive to coupled and locally varying structures. The B-only model, by contrast, gives the highest total hybrid win count, with 231 wins out of 400 comparisons, suggesting that cross-channel mixing produces the most frequent per-timestep advantage even when it does not minimize every mean metric.

Taken together, the 2020 results show that the hybrid advantage is not tied to a single dominant ansatz. At low channel count, the useful signal is weak, although the B-only $n_{ch}=3$ result already suggests that cross-channel bottleneck interactions are more effective than isolated within-channel processing. At $n_{ch}=9$, the variants become complementary: A-only is strongest for $u10m$, A+B is strongest for $v10m$, and B-only is strongest in total win count. This pattern suggests that different quantum circuit structures emphasize different aspects of the wind prediction task rather than providing a uniform improvement across all metrics.

\begin{table*}[t]
    \centering
    \caption{Summary of MAE, CRPS, and total per-timestep hybrid win counts for the classical baseline and the main hybrid configurations evaluated on the 2020 validation set. Lower MAE and CRPS indicate better performance. In the 9-channel regime, the hybrid variants show complementary strengths: A-only achieves the best MAE and CRPS for $u10m$, A+B achieves the best MAE and CRPS for $v10m$, and B-only obtains the highest total hybrid win count.}
    \label{tab:mae_crps_summary}
    \resizebox{\textwidth}{!}{%
    \begin{tabular}{lccccc}
        \toprule
        Model (2020) & MAE-$u10m$ & MAE-$v10m$ & CRPS-$u10m$ & CRPS-$v10m$ & Total Hybrid Wins \\
        \midrule
        Classical Corrdiff & 0.9658 $\pm$ 0.5473 & 1.0102 $\pm$ 0.5520 & 0.6999 $\pm$ 0.4131 & 0.7348 $\pm$ 0.4086 & -- \\
        \texttt{n\_circuit=1}($n_{ch}=3$), A+B & 0.9659 $\pm$ 0.5515 & 1.0170 $\pm$ 0.5677 & 0.7103 $\pm$ 0.4412 & 0.7504 $\pm$ 0.4465 & 182/400 (45.5\%) \\
        \texttt{n\_circuit=1}($n_{ch}=1$), A & 0.9653 $\pm$ 0.5520 & 1.0142 $\pm$ 0.5649 & 0.7122 $\pm$ 0.4448 & 0.7493 $\pm$ 0.4451 & 186/400 (46.5\%) \\
        \texttt{n\_circuit=3}($n_{ch}=3$), A & 0.9663 $\pm$ 0.5557 & 1.0140 $\pm$ 0.5650 & 0.7105 $\pm$ 0.4466 & 0.7471 $\pm$ 0.4392 & 193/400 (48.2\%) \\
        \texttt{n\_circuit=1}($n_{ch}=3$), B & 0.9561 $\pm$ 0.5541 & 1.0068 $\pm$ 0.5625 & 0.7089 $\pm$ 0.4521 & 0.7432 $\pm$ 0.4369 & 200/400 (50.0\%) \\
        \texttt{n\_circuit=3}($n_{ch}=9$), A+B & 0.9369 $\pm$ 0.5438 & \textbf{0.9979} $\pm$ 0.5219 & 0.6840 $\pm$ 0.4149 & \textbf{0.7269} $\pm$ 0.3830 & 215/400 (53.8\%) \\
        \texttt{n\_circuit=9}($n_{ch}=9$), A & \textbf{0.9250} $\pm$ 0.5566 & 0.9981 $\pm$ 0.5312 & \textbf{0.6783} $\pm$ 0.4377 & 0.7313 $\pm$ 0.4043 & 223/400 (55.8\%) \\
        \texttt{n\_circuit=3}($n_{ch}=9$), B & 0.9258 $\pm$ 0.5475 & 1.0009 $\pm$ 0.5623 & 0.6846 $\pm$ 0.4474 & 0.7396 $\pm$ 0.4398 & \textbf{231/400 (57.8\%)} \\
        \bottomrule
    \end{tabular}%
    }
\end{table*}

\subsection{Spectra and Distributions}
\begin{figure}[h!]
    \centering
    \includegraphics[width=0.99\linewidth]{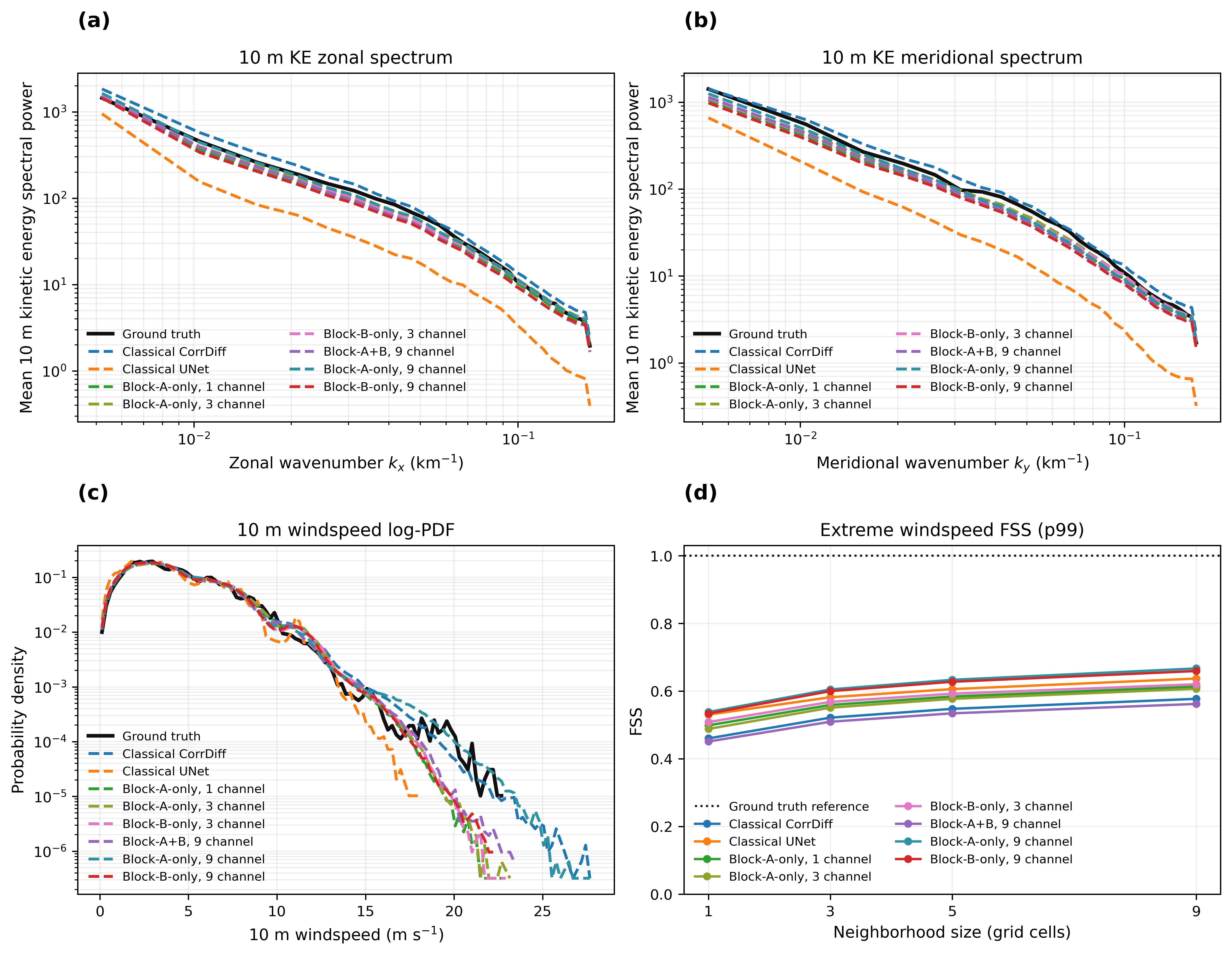}
    \caption{Structural realism and extreme-event localization diagnostics on the 2020 validation set. (a,b) Directional 10\,m kinetic-energy spectra along the zonal and meridional directions. Low wavenumbers correspond to broad flow structures, while high wavenumbers correspond to smaller-scale variability. The deterministic UNet is strongly underpowered, whereas CorrDiff restores the missing multiscale variance but is slightly over-energetic in parts of the low-to-intermediate wavenumber range. The hybrid variants preserve the CorrDiff-like spectral slope. (c) The 10\,m windspeed log-PDF shows that UNet under-represents the high-wind tail, CorrDiff produces the broadest far tail, and the hybrid variants occupy an intermediate regime. (d) The p99 FSS evaluates localization of the most extreme 1\% wind-speed events, with larger neighborhoods allowing larger spatial tolerance. The 9-channel hybrid variants recover part of the localization skill lost by classical CorrDiff, with Block-A-only and Block-B-only exceeding both CorrDiff and UNet across the tested neighborhoods.}
    \label{fig:structural_realism}
\end{figure}

\begin{figure*}[t]
    \centering
    \includegraphics[width=0.7\linewidth]{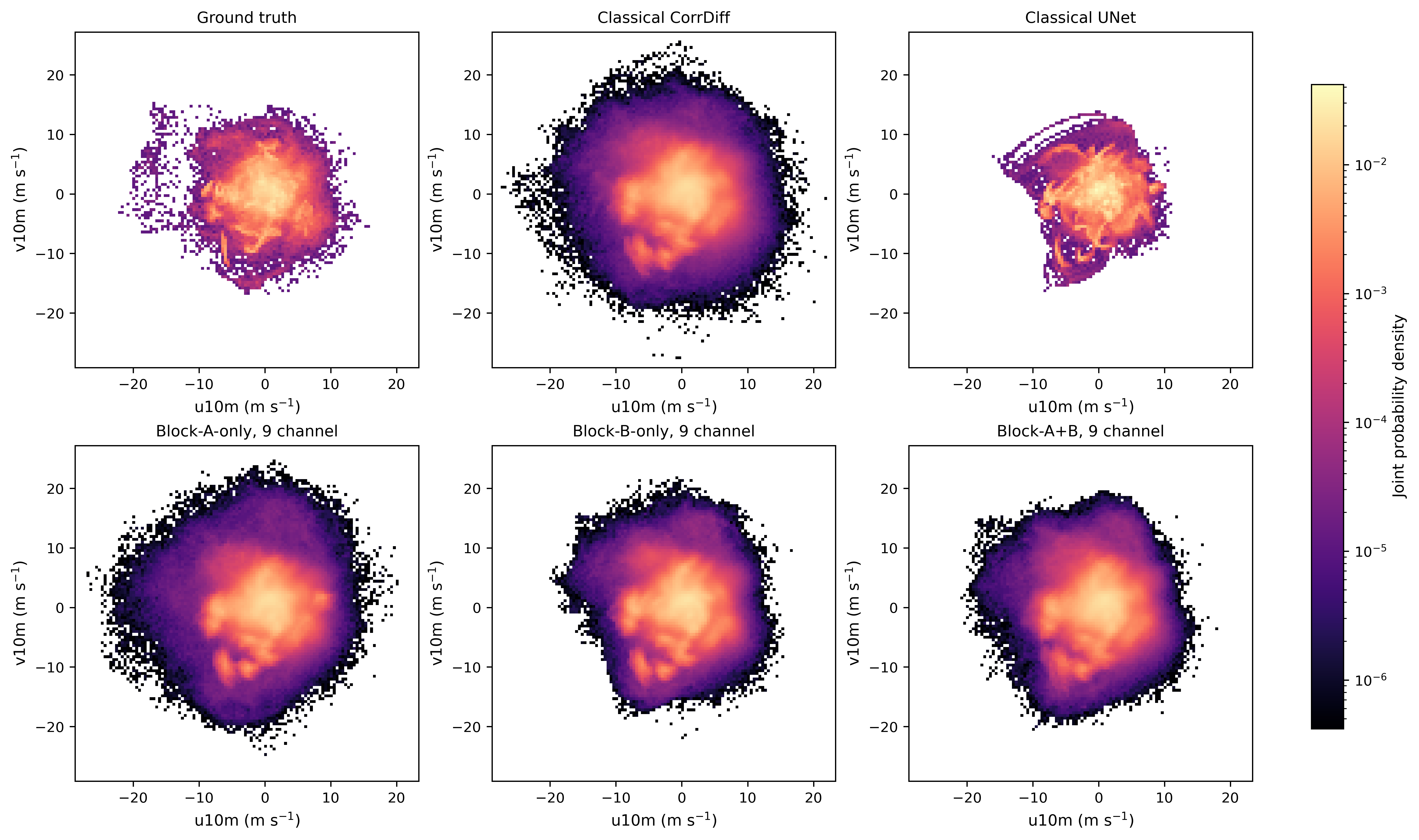}
    \caption{Joint probability density of the horizontal 10\,m wind components $(u10m,v10m)$ on the 2020 validation set. This diagnostic evaluates multivariate wind consistency: matching the one-dimensional marginals of $u10m$ and $v10m$ does not guarantee a realistic wind-vector distribution. The deterministic UNet produces a compressed support, reflecting regression toward smoother and less variable wind vectors. Classical CorrDiff broadens the support substantially but yields a more diffuse distribution than the ground truth. Among the hybrid variants, Block-A-only with 9 channels produces the broadest hybrid support, consistent with its stronger variance and high-wind tail retention, but it remains relatively diffuse. Block-B-only with 9 channels gives a more compact and coherent joint density while retaining broader support than UNet, indicating better organization of the coupled wind components. The full Block-A+B variant behaves as a mixed case and is not uniformly superior.}
    \label{fig:uv_joint_pdf}
\end{figure*}

Pointwise MAE and CRPS establish the primary predictive comparison, but they do not show whether the hybrid bottleneck preserves the multiscale and multivariate structure of the generated wind fields. We therefore complement the score-based validation with the structural diagnostics defined in Section~\ref{sec:evaluation_metrics}: directional 10\,m kinetic-energy spectra, the 10\,m windspeed log-PDF, p99 extreme-windspeed FSS, and the joint $(u10m,v10m)$ density. The first three are grouped in figure~\ref{fig:structural_realism}, while the joint distribution is shown separately in figure~\ref{fig:uv_joint_pdf} because it plays a distinct role as a multivariate consistency diagnostic.

The directional kinetic-energy spectra in fig. \ref{fig:structural_realism}(a) and (b) show the distribution of wind-field variability across spatial scales. Low wavenumbers correspond to broad flow structures, whereas high wavenumbers correspond to smaller-scale detail. The deterministic UNet strongly underestimates kinetic-energy power across the spectrum, reflecting oversmoothing. Whereas Classical CorrDiff restores much of this missing variability, but slightly exceeds the ground-truth power over parts of the low-to-intermediate wavenumber range. The hybrid model variants preserve the CorrDiff-like spectral shape while generally reducing this excess power, placing them between the underpowered UNet and the more energetic CorrDiff baseline. Among them, the 9-channel A-only model retains the most kinetic-energy power and remains closest to CorrDiff, while the B-only and A+B 9-channel models are marginally more damped. This suggests that intra-channel spatial processing better preserves wind-field variance, whereas cross-channel mixing acts more like a regularizing modification of the CorrDiff spectrum.

The windspeed log-PDF (figure~\ref{fig:structural_realism}(c)) exhibits the same distribution pattern. The UNet fails to capture the high-wind tail, while classical CorrDiff produces a longer and more oscillatory far tail than the ground truth. The hybrid model variants occupy an intermediate regime: they retain a broader tail than the UNet, but avoid part of the excessive tail extension of classical CorrDiff. Thus, the hybrid models do not simply reproduce the classical CorrDiff distribution; they modify the diffusion output in a controlled way while preserving its broader support relative to the deterministic baseline.

The p99 FSS (figure~\ref{fig:structural_realism}(d)) measures localization of the most extreme 1\% wind-speed events. Larger neighborhood sizes allow larger spatial tolerance, so increasing FSS with neighborhood size indicates that models capture the approximate event region even when exact pixel-level alignment is imperfect. Classical CorrDiff performs worse than the UNet in this metric, which is consistent with the underlying mechanism of stochastic diffusion samples having more spatial variability in the placement of extremes. The 9-channel hybrid model variants recover part of this localization skill, with A-only and B-only exceeding both classical CorrDiff and UNet across the tested neighborhoods. The full A+B model remains lower, indicating that combining both transformations does not automatically improve event placement.

The joint $(u10m,v10m)$ density in fig. \ref{fig:uv_joint_pdf} evaluates multivariate wind consistency. The UNet produces a compressed support, while classical CorrDiff broadens the distribution but appears more diffuse than the ground truth. A-only 9ch produces the broadest hybrid support, consistent with its stronger variance and high-wind tail retention, but it is also relatively diffuse. B-only 9ch gives a more compact and coherent joint density while retaining broader support than the UNet, suggesting that cross-channel processing better organizes the relation between the two wind components. The full A+B variant behaves as a mixed case and is not uniformly superior. Overall, A-only 9ch is strongest at preserving wind-field variance, B-only 9ch is strongest for extreme localization and compact multivariate structure, and A+B provides an intermediate combination rather than a dominant solution.

Taken together, these structural diagnostics support a coherent interpretation of the 2020 ID results. The hybrid bottleneck does not merely improve pointwise scores at the expense of physical realism. Instead, the best-performing configurations preserve the principal spectral and distributional signatures of CorrDiff while producing measurable, though still moderate, changes in extreme-event localization and multivariate wind structure.

\subsection{Robustness to simulated backend noise}

\begin{figure}
    \centering
    \includegraphics[width=0.95\linewidth]{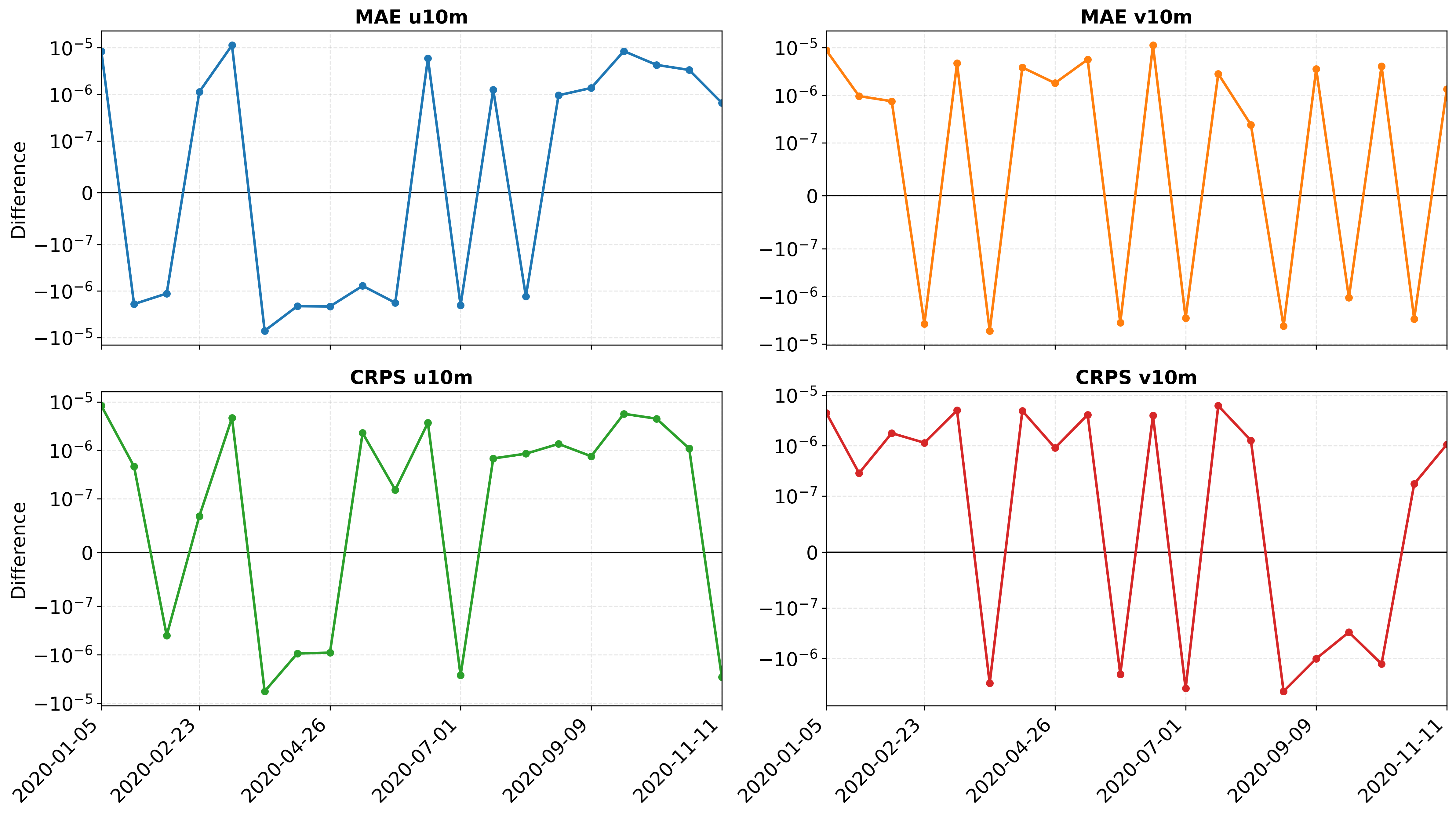}
    \caption{Differences in MAE and CRPS between predictions obtained with the IBM \texttt{fakefez} backend and the noiseless statevector simulator for the 10\,m wind components u10m and v10m. Differences are computed as \texttt{(fakefez - statevector)} for $\texttt{n\_circuits}=1$ using the Block-B-only HQConv ansatz, over 20 evenly spaced verification times from 2020 and 16 ensemble members. The differences remain on the order of $10^{-6}$, indicating that simulated device noise has a negligible effect on the evaluated error metrics at the present circuit depth and model configuration.}
    \label{fig:fakefez_minus_noiseless_diff}
\end{figure}

We assess simulated backend-noise sensitivity by comparing hybrid inference with Pennylane's noiseless \texttt{default.qubit} simulator and the IBM \texttt{fakefez} noisy backend emulator, which includes device-level gate and readout noise. The comparison is performed on 20 evenly spaced 2020 verification times, using 16 ensemble members per time; MAE is computed from the ensemble mean, and CRPS from the generated ensemble. As shown in figure~\ref{fig:fakefez_minus_noiseless_diff}, the differences between noisy and noiseless simulation remain on the order of $10^{-6}$ for both $u10m$ and $v10m$ across MAE and CRPS. These differences are negligible compared with the absolute metric values and with the natural variability across verification times. Thus, at the present circuit depth and qubit count, simulated device noise does not materially alter the observed validation trends. The larger deployment gap discussed below is therefore mainly associated with real-hardware execution constraints rather than the backend-noise level represented by \texttt{fakefez}.

\subsection{Real-hardware deployment under qubit constraints}

In this section, the results on whether the hybrid CorrDiff model retains meaningful behaviour when the quantum bottleneck is executed on a real QPU are presented and evaluated. Since the JIQCER-5 device provides only five physical qubits, the real QPU runs are restricted to the minimal compatible configuration: Block-A-only ansatz, which encodes one quantum-processed bottleneck channel with a 4-qubit circuit. We therefore evaluate a reduced inference setting with \texttt{n\_circuits=1} and 4 ensemble members on four representative 2020 dates from spring, summer, autumn, and winter. For reference, the classical CorrDiff baseline, the noiseless quantum simulation, and the noisy \texttt{fakefez} simulation are evaluated under a closely matched reduced setting. As a testing dataset, 4 different dates across the year (one for each season) were chosen.

\begin{figure}
    \centering
    \includegraphics[width=\linewidth]{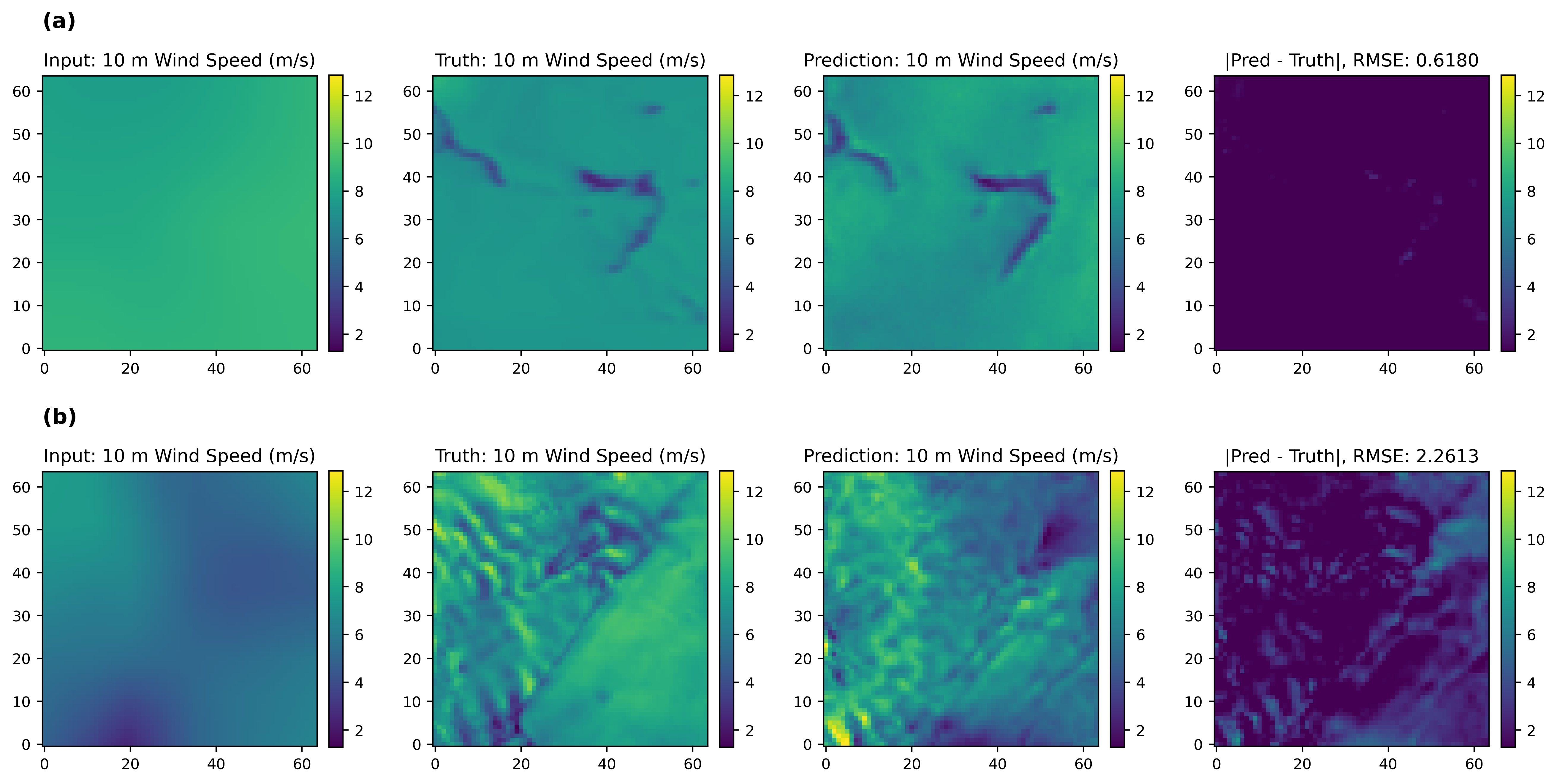}
    \caption{Wind map prediction for the datestamp 14.04.2020 (a) and 01.07.2020 (b) with 1 channel composed of 4 qubits.}
    \label{fig:overall_plots_n_channels_1}
\end{figure}

The representative real-hardware map comparisons in fig. \ref{fig:overall_plots_n_channels_1} present one favorable case and one stress case for the real device experiments. The 14.04.2020 map has the lowest wind-speed RMSE among the four real-hardware map evaluations and corresponds to a relatively smooth target field. In this case, the JIQCER-5 prediction preserves the dominant large-scale structure and remains close to the truth. The 01.07.2020 map has the highest wind-speed RMSE among the four displayed hardware cases and contains stronger localized gradients and filamentary structures. It therefore illustrates the regime in which real-device execution most visibly loses fine-scale fidelity. Together, these two maps show that the deployed model does not collapse on the real QPU, but that its errors concentrate in the reconstruction of small-scale structure rather than modelling the coarse flow.

For reference, the classical CorrDiff baseline, the noiseless statevector simulation, and the noisy IBM \texttt{fakefez} backend simulation are evaluated under a closely matched reduced setting with 4 ensemble members and 20 verification timesteps distributed across 2020. The real-device experiments are more restricted due to the execution overhead and use the minimal hardware-compatible configuration, namely one quantum-processed bottleneck channel represented by a 4-qubit circuit, 4 ensemble members, and four representative verification dates: 05.01.2020, 14.04.2020, 01.07.2020, and 27.09.2020. This setup is therefore a deployment-oriented feasibility test rather than a full-scale performance benchmark.

\begin{figure}[h!]
    \centering
    \includegraphics[width=\linewidth]{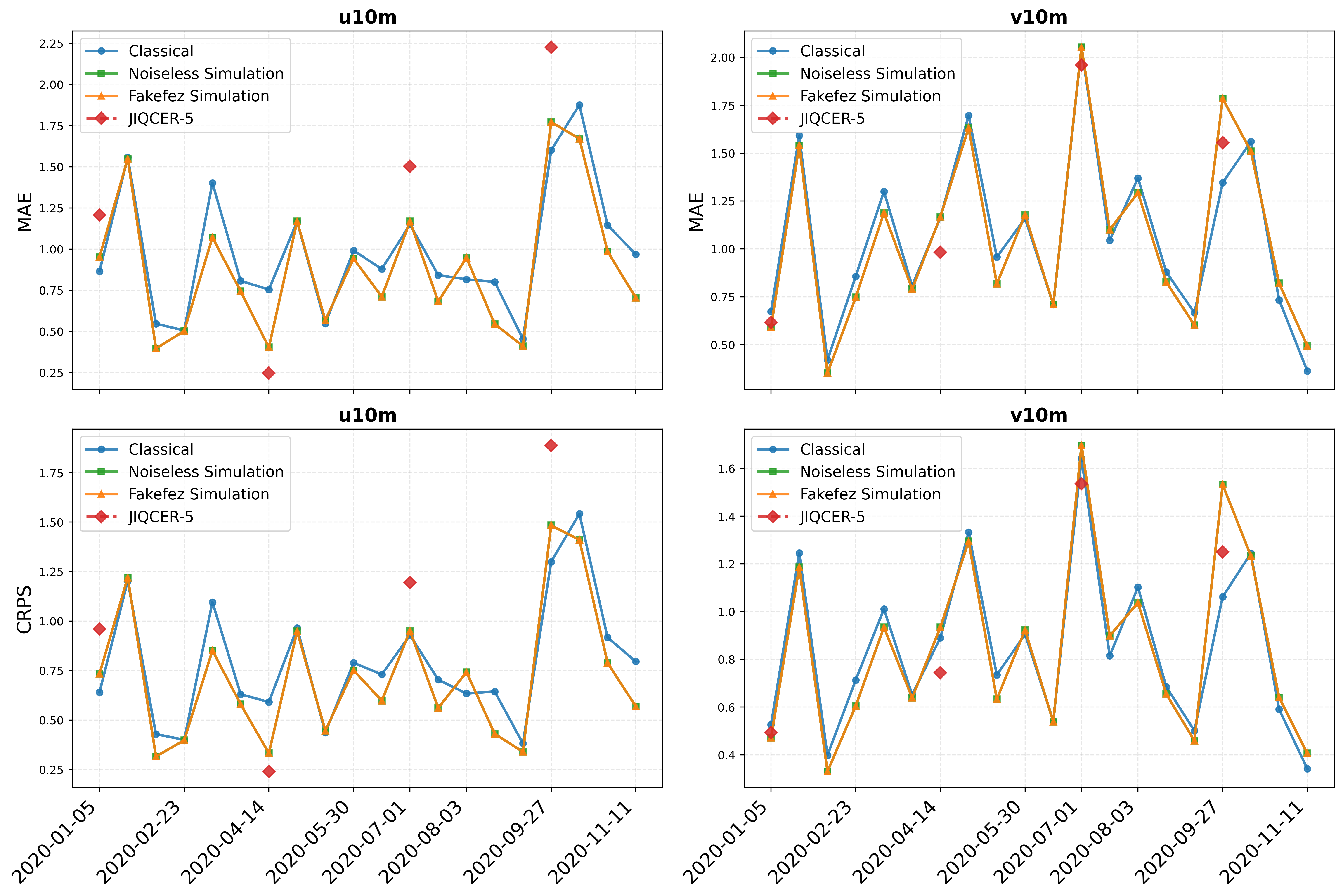}
    \caption{MAE and CRPS comparison for u10m and v10m between the classical CorrDiff baseline, noiseless quantum simulation, fakefez noisy simulation, and real JIQCER-5 hardware on verification dates in 2020.}
    \label{fig:mae_crps_multimodel}
\end{figure}

The MAE and CRPS profiles in figure~\ref{fig:mae_crps_multimodel} show that the real-hardware behavior is mixed rather than uniformly degraded. The noiseless and \texttt{fakefez} curves almost overlap, confirming that simulated backend noise has a negligible impact at this circuit scale. The JIQCER-5 points, however, deviate more visibly from the simulation curves, indicating that real-device conditions introduce additional variability beyond the emulator-level noise model.

For $u10m$, the real-hardware results vary strongly across the four dates. The 14.04.2020 case is substantially better than the classical and simulated references, whereas 01.07.2020 and especially 27.09.2020 show clear degradation. Hence, this component exposes the largest hardware sensitivity in our deployment setting. For $v10m$, the real-hardware scores are more competitive: the JIQCER-5 points remain close to the classical and simulated references on several dates, including 01.07.2020 and 27.09.2020, although the four-date sample is too small to claim a robust advantage.

Taken together, the maps and score profiles indicate that real-hardware execution can retain meaningful meteorological structure in simple cases, but remains less stable than noiseless or noisy simulation backends. The main visible limitation is the loss of fine-scale fidelity under more complex wind patterns, while the coarse flow structure is often preserved. This supports the interpretation that present Noisy Intermediate-Scale Quantum (NISQ) device deployment is feasible as a proof of concept, but is still constrained by qubit count, circuit fidelity, and execution overhead.

\subsection{OOD stress test and motivation for mitigation}

The previous 2020 validation results on simulation and the real device establish that the hybrid bottleneck can yield repeated in-distribution gains without disrupting the main structural statistics of CorrDiff. The natural next question is whether these gains persist under temporal distribution shifts. To test this, we evaluate the main hybrid models in an OOD setting, where training uses data from  earlier years (2018-2020) and evaluation is performed on later-year data (2021).

\begin{table*}[h!]
    \centering
    \caption{Summary of MAE, CRPS, and total per-timestep hybrid win counts for the classical baseline and the hybrid HQConv configurations evaluated on the 2021 out-of-distribution (OOD) test set. Lower MAE and CRPS indicate better performance. In contrast to the 2020 in-distribution results, the classical baseline remains best on average for all four metrics, while the Block-B-only configuration with $ch=3$ attains the highest hybrid win count, indicating that OOD gains are less stable and more configuration-dependent.}
    \label{tab:mae_crps_summary_2021}
    \resizebox{\textwidth}{!}{%
    \begin{tabular}{lccccc}
        \toprule
        Model (2021) & MAE-$u10m$ & MAE-$v10m$ & CRPS-$u10m$ & CRPS-$v10m$ & Total Hybrid Wins \\
        \midrule
        Classical Corrdiff & \textbf{0.9733} $\pm$ 0.5299 & \textbf{0.9368} $\pm$ 0.4274 & \textbf{0.7259} $\pm$ 0.4375 & \textbf{0.6917} $\pm$ 0.3459 & -- \\
        $n_{ch}=1$, A & 0.9950 $\pm$ 0.5353 & 0.9854 $\pm$ 0.4938 & 0.7282 $\pm$ 0.4119 & 0.7271 $\pm$ 0.3735 & 163/400 (40.8\%) \\
        $n_{ch}=3$, A & 0.9935 $\pm$ 0.5411 & 0.9595 $\pm$ 0.4740 & 0.7301 $\pm$ 0.4293 & 0.7074 $\pm$ 0.3719 & 190/400 (47.5\%) \\
        $n_{ch}=3$, B & 0.9890 $\pm$ 0.5344 & 0.9565 $\pm$ 0.4686 & 0.7288 $\pm$ 0.4288 & 0.7064 $\pm$ 0.3756 & \textbf{204/400 (51.0\%)} \\
        $n_{ch}=9$, A & 0.9871 $\pm$ 0.5345 & 0.9550 $\pm$ 0.4650 & 0.7275 $\pm$ 0.4277 & 0.7030 $\pm$ 0.3680 & 185/400 (46.2\%) \\
        $n_{ch}=9$, B & 0.9914 $\pm$ 0.5398 & 0.9699 $\pm$ 0.4876 & 0.7289 $\pm$ 0.4304 & 0.7158 $\pm$ 0.3826 & 190/400 (47.5\%) \\
        \bottomrule
    \end{tabular}%
    }
\end{table*}

The OOD results are summarized in table~\ref{tab:mae_crps_summary_2021}. In contrast to the 2020 ID setting, the hybrid gains do not transfer uniformly under temporal shift. Some configurations that were favorable in the 2020 validation setting no longer outperform the classical CorrDiff baseline under the 2021 OOD test, particularly in terms of MAE. CRPS generalizes somewhat better than MAE across the hybrid variants, but the classical baseline still achieves the lowest average CRPS values. This does not negate the ID findings. Rather, it clarifies their scope: the HQConv bottleneck can improve the classical diffusion model under matched validation conditions, but these gains remain sensitive to training dynamics and to the degree of distribution shift.

This OOD behavior is therefore best interpreted not as a contradiction, but as a diagnosis. It indicates that bottleneck-level hybridization introduces additional optimization sensitivity, which in turn motivates mitigation strategies such as separate learning-rate scheduling and fusion regularization. In this sense, the OOD experiment serves as a transition from the present unmitigated HQConv study to the stabilization strategies developed in the subsequent part of the work.

\section{Conclusion}

In this project, we presented an approach for combining classical and quantum computing into a hybrid quantum-classical diffusion model architecture. This is a very flexible approach, as it allows encoding as many internal features as possible within a quantum circuit, from a minimum of 4 (namely one latent channel), each feature requiring one qubit. We evaluated hardware deployment on JIQCER-5 (IQM Spark) for the case of 4 features, and evaluated ideal and simulated noisy behavior for up to 36 features with three 12-qubit circuits. The main findings can be summarized as follows:
\begin{itemize}
    \item The presented hybrid model maintains a competitive performance on the evaluated downscaling benchmark, with a slight observed average improvement over the fully classical counterpart.
    \item Given the lower number of parameters in the hybrid UNet block compared to the fully classical one, the hybrid approach can also be seen as a parameter reduction approach with no degradation in performance.
    \item Experience with real-hardware implementation argues against any definitive performance degradation, which is still more visible with experiments on JIQCER-5, although still preserving broad wind patterns in the generated images.
    \item However, at the tested scale (i.e., 4 qubits), the time overhead becomes prohibitive for large-scale experimentation, increasing the runtime scale by an order of magnitude, even excluding overheads introduced by job scheduling and queuing. Nevertheless, the scaling of the proposed circuit and the gate operation times of superconducting quantum systems indicate near-constant runtime scaling for increasing number of qubits.
\end{itemize}

The results support a constrained but meaningful role for quantum bottleneck layers in probabilistic weather downscaling. The study demonstrates that quantum circuit processing latent-space information can produce measurable and physically interpretable changes in a competitive diffusion downscaling model, while the OOD generalization gap points to a clear mitigation path through more stable quantum-classical optimization and controlled feature fusion, such as separate learning-rate schedules and learnable fusion gates. This interpretation is consistent with known trainability challenges in variational quantum models, where optimization can be sensitive to initialization, circuit structure, and gradient concentration \cite{mcclean2018barren,grant2019initialization,cerezo2021cost}.

Generative modeling with hybrid quantum-classical methods is a relatively new and promising direction for practical quantum machine learning, given its flexibility in terms of hardware requirements. Nevertheless, several hurdles have to be overcome for it to become a core module of operational Earth observation tools within Digital Twin Earth. Having dedicated access is the first necessary step. Nevertheless, even an on-premises machine with minimal queuing time still poses a significant overhead, as observed in the experiments. A tighter hardware and software integration between classical and quantum resources would reduce this overhead, with a large overall impact on repeated calls. This is still an active research direction on which several actors are working, and for which significant results are expected to be seen in a few years. High qubit count machines would both reduce quantum machine calls and reduce model parameters, drastically improving the overall runtime. This can lead to comparable or better performance provided that error rates are sufficiently low, a positive example being the experiments with the simulated IBM machines. The high cost of installing and maintaining such machines is also a factor that hinders widespread adoption. In addition to hardware limitations, research on model development can also have a substantial impact on the adoption roadmap, setting different hardware requirements for the same task.

\bibliographystyle{ieeetr}
\bibliography{bibliography.bib}

@article{ho2020denoising,
  title={Denoising diffusion probabilistic models},
  author={Ho, Jonathan and Jain, Ajay and Abbeel, Pieter},
  journal={Advances in neural information processing systems},
  volume={33},
  pages={6840--6851},
  year={2020}
}

@article{mardani2025residual,
  title={Residual corrective diffusion modeling for km-scale atmospheric downscaling},
  author={Mardani, Morteza and Brenowitz, Noah and Cohen, Yair and Pathak, Jaideep and Chen, Chieh-Yu and Liu, Cheng-Chin and Vahdat, Arash and Nabian, Mohammad Amin and Ge, Tao and Subramaniam, Akshay and others},
  journal={Communications Earth \& Environment},
  volume={6},
  number={1},
  pages={124},
  year={2025},
  publisher={Nature Publishing Group UK London}
}

@article{Skamarock2014atmospheric,
      author = "William C. Skamarock and Sang-Hun Park and Joseph B. Klemp and Chris Snyder",
      title = "Atmospheric Kinetic Energy Spectra from Global High-Resolution Nonhydrostatic Simulations",
      journal = "Journal of the Atmospheric Sciences",
      year = "2014",
      publisher = "American Meteorological Society",
      address = "Boston MA, USA",
      volume = "71",
      number = "11",
      doi = "10.1175/JAS-D-14-0114.1",
      pages = "4369 - 4381",
      url = "https://journals.ametsoc.org/view/journals/atsc/71/11/jas-d-14-0114.1.xml"
}

@article{roberts2008scale,
      author = "Nigel M. Roberts and Humphrey W. Lean",
      title = "Scale-Selective Verification of Rainfall Accumulations from High-Resolution Forecasts of Convective Events",
      journal = "Monthly Weather Review",
      year = "2008",
      publisher = "American Meteorological Society",
      address = "Boston MA, USA",
      volume = "136",
      number = "1",
      doi = "10.1175/2007MWR2123.1",
      pages=      "78 - 97",
      url = "https://journals.ametsoc.org/view/journals/mwre/136/1/2007mwr2123.1.xml"
}

@article{cannon2016multivariate,
      author = "Alex J. Cannon",
      title = "Multivariate Bias Correction of Climate Model Output: Matching Marginal Distributions and Intervariable Dependence Structure",
      journal = "Journal of Climate",
      year = "2016",
      publisher = "American Meteorological Society",
      address = "Boston MA, USA",
      volume = "29",
      number = "19",
      doi = "10.1175/JCLI-D-15-0679.1",
      pages=      "7045 - 7064",
      url = "https://journals.ametsoc.org/view/journals/clim/29/19/jcli-d-15-0679.1.xml"
}

@article{wang2024multivariate,
author={Wang, Fang
and Tian, Di},
title={Multivariate bias correction and downscaling of climate models with trend-preserving deep learning},
journal={Climate Dynamics},
year={2024},
month={Oct},
day={01},
volume={62},
number={10},
pages={9651-9672},
issn={1432-0894},
doi={10.1007/s00382-024-07406-9},
url={https://doi.org/10.1007/s00382-024-07406-9}
}

@article{de2024towards,
  title={Towards efficient quantum hybrid diffusion models},
  author={De Falco, Francesca and Ceschini, Andrea and Sebastianelli, Alessandro and Saux, Bertrand Le and Panella, Massimo},
  journal={arXiv preprint arXiv:2402.16147},
  year={2024}
}

@article{benestad2008,
author = {Benestad, R. and Hanssen-Bauer, Inger and Chen, Deliang},
year = {2008},
month = {09},
pages = {},
title = {Empirical-Statistical Downscaling},
isbn = {978-981-281-912-3},
journal = {World Scientific Editions},
doi = {10.1142/6908}
}

@article{stoner2013,
author = {Stoner, Anne Marie and Hayhoe, Katharine and Yang, Xiaohui and Wuebbles, Donald},
year = {2013},
month = {09},
pages = {},
title = {An Asynchronous Regional Regression Model for Statistical Downscaling of Daily Climate Variables},
volume = {33},
journal = {International Journal of Climatology},
doi = {10.1002/joc.3603}
}

@article{Thrasher2022,
  title = {NASA Global Daily Downscaled Projections,  CMIP6},
  volume = {9},
  ISSN = {2052-4463},
  url = {http://dx.doi.org/10.1038/s41597-022-01393-4},
  DOI = {10.1038/s41597-022-01393-4},
  number = {1},
  journal = {Scientific Data},
  publisher = {Springer Science and Business Media LLC},
  author = {Thrasher,  Bridget and Wang,  Weile and Michaelis,  Andrew and Melton,  Forrest and Lee,  Tsengdar and Nemani,  Ramakrishna},
  year = {2022},
  month = jun 
}

@article{bergholm2018pennylane,
  title={Pennylane: Automatic differentiation of hybrid quantum-classical computations},
  author={Bergholm, Ville and Izaac, Josh and Schuld, Maria and Gogolin, Christian and Ahmed, Shahnawaz and Ajith, Vishnu and Alam, M Sohaib and Alonso-Linaje, Guillermo and AkashNarayanan, B and Asadi, Ali and others},
  journal={arXiv preprint arXiv:1811.04968},
  year={2018}
}

@article{schuld2019evaluating,
  title={Evaluating analytic gradients on quantum hardware},
  author={Schuld, Maria and Bergholm, Ville and Gogolin, Christian and Izaac, Josh and Killoran, Nathan},
  journal={Physical Review A},
  volume={99},
  number={3},
  pages={032331},
  year={2019},
  publisher={APS}
}

@article{gneiting2007strictly,
  title={Strictly proper scoring rules, prediction, and estimation},
  author={Gneiting, Tilmann and Raftery, Adrian E},
  journal={Journal of the American Statistical Association},
  volume={102},
  number={477},
  pages={359--378},
  year={2007},
  month={mar}
}

@article{Iorio2004,
author = {J.P, Iorio and Duffy, Philip and Govindasamy, Balasubramanian and Thompson, S. and Randall, D.},
year = {2004},
month = {01},
pages = {243-258},
title = {Effect of model resolution and subgrif scale physics on the simulation of precipitation in the continental United States},
volume = {23},
journal = {Climate Dynamics},
doi = {10.1007/s00382-004-0440-y}
}

@article{SACHINDRA2018240,
title = {Statistical downscaling of precipitation using machine learning techniques},
journal = {Atmospheric Research},
volume = {212},
pages = {240-258},
year = {2018},
issn = {0169-8095},
doi = {https://doi.org/10.1016/j.atmosres.2018.05.022},
url = {https://www.sciencedirect.com/science/article/pii/S0169809517310141},
author = {D.A. Sachindra and K. Ahmed and Md. Mamunur Rashid and S. Shahid and B.J.C. Perera}
}

@article{Rashid2015,
author = {Rashid, Md Mamunur and Beecham, Simon and Chowdhury, Rezaul},
year = {2015},
month = {04},
pages = {},
title = {Statistical downscaling of rainfall: a non-stationary and multi-resolution approach},
volume = {124},
journal = {Theoretical and Applied Climatology},
doi = {10.1007/s00704-015-1465-3}
}

@article{Sachindra2015,
  title = {Impact of climate change on urban heat island effect and extreme temperatures: a case‐study},
  volume = {142},
  ISSN = {1477-870X},
  url = {http://dx.doi.org/10.1002/qj.2642},
  DOI = {10.1002/qj.2642},
  number = {694},
  journal = {Quarterly Journal of the Royal Meteorological Society},
  publisher = {Wiley},
  author = {Sachindra,  D. A. and Ng,  A. W. M. and Muthukumaran,  S. and Perera,  B. J. C.},
  year = {2015},
  month = Sept,
  pages = {172–186}
}

@article{yang2023diffusion,
  title={Diffusion models: A comprehensive survey of methods and applications},
  author={Yang, Ling and Zhang, Zhilong and Song, Yang and Hong, Shenda and Xu, Runsheng and Zhao, Yue and Zhang, Wentao and Cui, Bin and Yang, Ming-Hsuan},
  journal={ACM computing surveys},
  volume={56},
  number={4},
  pages={1--39},
  year={2023},
  publisher={ACM New York, NY, USA}
}

@article { gutowski2020,
      author = "W. J. Gutowski and P. A. Ullrich and A. Hall and L. R. Leung and T. A. O’Brien and C. M. Patricola-DiRosario and R. W. Arritt and M. S. Bukovsky and K. V. Calvin and Z. Feng and A. D. Jones and G. J. Kooperman and E. Monier and M. S. Pritchard and S. C. Pryor and Y. Qian and A. M. Rhoades and A. F. Roberts and K. Sakaguchi and N. Urban and C. Zarzycki",
      title = "The Ongoing Need for High-Resolution Regional Climate Models: Process Understanding and Stakeholder Information",
      journal = "Bulletin of the American Meteorological Society",
      year = "2020",
      publisher = "American Meteorological Society",
      address = "Boston MA, USA",
      volume = "101",
      number = "5",
      doi = "10.1175/BAMS-D-19-0113.1",
      pages=      "E664 - E683",
      url = "https://journals.ametsoc.org/view/journals/bams/101/5/bams-d-19-0113.1.xml"
}

@inproceedings{Sachindra2011,
author = {Sachindra, D.A. and Huang, F and Barton, Andrew and Perera, B.},
year = {2011},
month = {01},
pages = {},
title = {Statistical Downscaling of General Circulation Model Outputs to Catchment Streamflows}
}

@article {Selz2015,
      author = "Tobias Selz and George C. Craig",
      title = "Upscale Error Growth in a High-Resolution Simulation of a Summertime Weather Event over Europe",
      journal = "Monthly Weather Review",
      year = "2015",
      publisher = "American Meteorological Society",
      address = "Boston MA, USA",
      volume = "143",
      number = "3",
      doi = "10.1175/MWR-D-14-00140.1",
      pages=      "813 - 827",
      url = "https://journals.ametsoc.org/view/journals/mwre/143/3/mwr-d-14-00140.1.xml"
}

@inproceedings{ronneberger2015u,
  title={U-net: Convolutional networks for biomedical image segmentation},
  author={Ronneberger, Olaf and Fischer, Philipp and Brox, Thomas},
  booktitle={International Conference on Medical image computing and computer-assisted intervention},
  pages={234--241},
  year={2015},
  organization={Springer}
}

@article{
Stengel2020,
author = {Karen Stengel  and Andrew Glaws  and Dylan Hettinger  and Ryan N. King },
title = {Adversarial super-resolution of climatological wind and solar data},
journal = {Proceedings of the National Academy of Sciences},
volume = {117},
number = {29},
pages = {16805-16815},
year = {2020},
doi = {10.1073/pnas.1918964117},
URL = {https://www.pnas.org/doi/abs/10.1073/pnas.1918964117},
eprint = {https://www.pnas.org/doi/pdf/10.1073/pnas.1918964117}}

@article{nguyen2023climax,
  title={Climax: A foundation model for weather and climate},
  author={Nguyen, Tung and Brandstetter, Johannes and Kapoor, Ashish and Gupta, Jayesh K and Grover, Aditya},
  journal={arXiv preprint arXiv:2301.10343},
  year={2023}
}

@ARTICLE{Leinonen2021,
  author={Leinonen, Jussi and Nerini, Daniele and Berne, Alexis},
  journal={IEEE Transactions on Geoscience and Remote Sensing}, 
  title={Stochastic Super-Resolution for Downscaling Time-Evolving Atmospheric Fields With a Generative Adversarial Network}, 
  year={2021},
  volume={59},
  number={9},
  pages={7211-7223},
  doi={10.1109/TGRS.2020.3032790}}

@article{Vosper2023,
author = {Vosper, Emily and Watson, Peter and Harris, Lucy and McRae, Andrew and Santos-Rodriguez, Raul and Aitchison, Laurence and Mitchell, Dann},
title = {Deep Learning for Downscaling Tropical Cyclone Rainfall to Hazard-Relevant Spatial Scales},
journal = {Journal of Geophysical Research: Atmospheres},
volume = {128},
number = {10},
pages = {e2022JD038163},
doi = {https://doi.org/10.1029/2022JD038163},
url = {https://agupubs.onlinelibrary.wiley.com/doi/abs/10.1029/2022JD038163},
eprint = {https://agupubs.onlinelibrary.wiley.com/doi/pdf/10.1029/2022JD038163},
note = {e2022JD038163 2022JD038163},
year = {2023}
}

@article{molinaro2026universal,
  title={Universal Diffusion-Based Probabilistic Downscaling},
  author={Molinaro, Roberto and Siegenheim, Niall and Martin, Henry and Frey, Mark and Poulsen, Niels and Seitz, Philipp and Gabler, Marvin Vincent},
  journal={arXiv preprint arXiv:2602.11893},
  year={2026}
}

@article{hatanaka2023diffusion,
  title={Diffusion models for high-resolution solar forecasts},
  author={Hatanaka, Yusuke and Glaser, Yannik and Galgon, Geoff and Torri, Giuseppe and Sadowski, Peter},
  journal={arXiv preprint arXiv:2302.00170},
  year={2023}
}

@article{mcclean2018barren,
author={McClean, Jarrod R.
and Boixo, Sergio
and Smelyanskiy, Vadim N.
and Babbush, Ryan
and Neven, Hartmut},
title={Barren plateaus in quantum neural network training landscapes},
journal={Nature Communications},
year={2018},
month={Nov},
day={16},
volume={9},
number={1},
pages={4812},
issn={2041-1723},
doi={10.1038/s41467-018-07090-4},
url={https://doi.org/10.1038/s41467-018-07090-4}
}

@article{grant2019initialization,
  doi = {10.22331/q-2019-12-09-214},
  url = {https://doi.org/10.22331/q-2019-12-09-214},
  title = {An initialization strategy for addressing barren plateaus in parametrized quantum circuits},
  author = {Grant, Edward and Wossnig, Leonard and Ostaszewski, Mateusz and Benedetti, Marcello},
  journal = {{Quantum}},
  issn = {2521-327X},
  publisher = {{Verein zur F{\"{o}}rderung des Open Access Publizierens in den Quantenwissenschaften}},
  volume = {3},
  pages = {214},
  month = dec,
  year = {2019}
}

@article{cerezo2021cost,
author={Cerezo, M.
and Sone, Akira
and Volkoff, Tyler
and Cincio, Lukasz
and Coles, Patrick J.},
title={Cost function dependent barren plateaus in shallow parametrized quantum circuits},
journal={Nature Communications},
year={2021},
month={Mar},
day={19},
volume={12},
number={1},
pages={1791},
issn={2041-1723},
doi={10.1038/s41467-021-21728-w},
url={https://doi.org/10.1038/s41467-021-21728-w}
}

\end{document}